\documentclass{article} 
\usepackage{arxiv,times}

\usepackage{amsmath,amsfonts,bm}

\def\eqref#1{equation~\ref{#1}}

\def\1{\bm{1}}

\DeclareMathAlphabet{\mathsfit}{\encodingdefault}{\sfdefault}{m}{sl}
\SetMathAlphabet{\mathsfit}{bold}{\encodingdefault}{\sfdefault}{bx}{n}

\usepackage{url}

\usepackage[utf8]{inputenc}
\definecolor{prompt}{RGB}{211, 248, 226}
\definecolor{repre}{RGB}{228, 193, 249}
\definecolor{mixture}{RGB}{237, 231, 177}
\definecolor{promptable}{RGB}{226, 255, 237}
\definecolor{repretable}{RGB}{238, 221, 248}
\definecolor{mixturetable}{RGB}{250, 246, 210}
\definecolor{mirrortable}{RGB}{185, 235, 205}
\definecolor{mixturetooltable}{RGB}{235, 222, 155}
\definecolor{training}{RGB}{166, 205, 239}
\definecolor{trainingtable}{RGB}{230, 242, 253}
\newcommand{\markTrain}{\textcolor{training!80!black}{$\blacklozenge$}}
\newcommand{\markTool}{\textcolor{prompt!80!black}{$\blacktriangle$}}
\newcommand{\markBase}{\textcolor{repre!80!black}{$\bullet$}}
\newcommand{\markFw}{\textcolor{mixture!80!black}{$\blacksquare$}}

\usepackage{graphicx}
\usepackage{booktabs}
\usepackage{multirow}
\usepackage[table]{xcolor}
\usepackage{array}
\usepackage{makecell}
\usepackage{caption}
\usepackage{tabularx}
\usepackage{adjustbox}
\usepackage{amssymb}   
\usepackage{xcolor}    
\usepackage{pifont}    
\usepackage{subcaption}
\usepackage{calc}
\usepackage{changes}

\usepackage{tcolorbox}
\tcbuselibrary{listings, breakable}
\usepackage{xcolor}
\usepackage{enumitem}
\definecolor{titlegray}{HTML}{4A4A4A}
\definecolor{bordergray}{HTML}{4A4A4A}

\newcommand{\ouragent}{EvoGuard}

\newcommand{\mechanismnameshort}{capability-aware dynamic orchestration}

\newcommand{\toolprofile}{tool profile}

\definecolor{linkblue}{rgb}{0.21,0.49,0.74}
\usepackage[pagebackref,breaklinks,colorlinks,allcolors=linkblue]{hyperref}
\usepackage{cleveref}

\title{\ouragent: An Extensible Agentic RL-based Framework for Practical and Evolving AI-Generated Image Detection}

\author{
Chenyang Zhu$^{1,2}$, Maorong Wang$^{2}$, Jun Liu$^{2}$, 
Ching-Chun Chang$^{2}$ \& Isao Echizen$^{2,1}$\\
\textsuperscript{1}The University of Tokyo \quad
\textsuperscript{2}National Institute of Informatics\\
Tokyo, Japan\\
\texttt{chenyangzhu\_public@163.com}\\
\texttt{\{maorong,csjunliu,ccchang,iechizen\}@nii.ac.jp}
}

\begin{document}

\maketitle

\begin{abstract}
The rapid proliferation of AI-Generated Images (AIGIs) poses severe misinformation risks, making AIGI detection critical yet challenging. 
Traditional detection paradigms mainly rely on low-level features, whereas recent research increasingly focuses on leveraging the general understanding ability of Multimodal Large Language Models (MLLMs) to achieve better generalization, yet it still suffers from limited extensibility and expensive data annotations.
Instead of building yet another detector, we recast AIGI detection as learned, reasoning-based evidence synthesis over a pool of heterogeneous off-the-shelf detectors, realized through EvoGuard, a novel agentic framework.
A capability-aware selection mechanism profiles each detector and gathers complementary evidence per sample; a dynamic orchestration mechanism then reasons over heterogeneous outputs across multiple rounds, cross-validating conflicting or low-confidence signals before concluding. 
This design exploits the complementary strengths among heterogeneous detectors, transcending the limits of any single model. 
Furthermore, optimized by a GRPO-based Agentic Reinforcement Learning algorithm using only low-cost binary labels, it eliminates the reliance on fine-grained annotations.  
Extensive experiments demonstrate that this learned reasoning paradigm outperforms single-detector and static ensembling, achieving SOTA accuracy while mitigating the bias between positive and negative samples.
More importantly, it allows the plug-and-play integration of new detectors to boost overall performance in a train-free manner, offering a highly practical, long-term solution to ever-evolving AIGI threats. 
Source code will be made publicly available.
\end{abstract}

\begin{figure*}[h]
\centering
\includegraphics[scale=0.17]{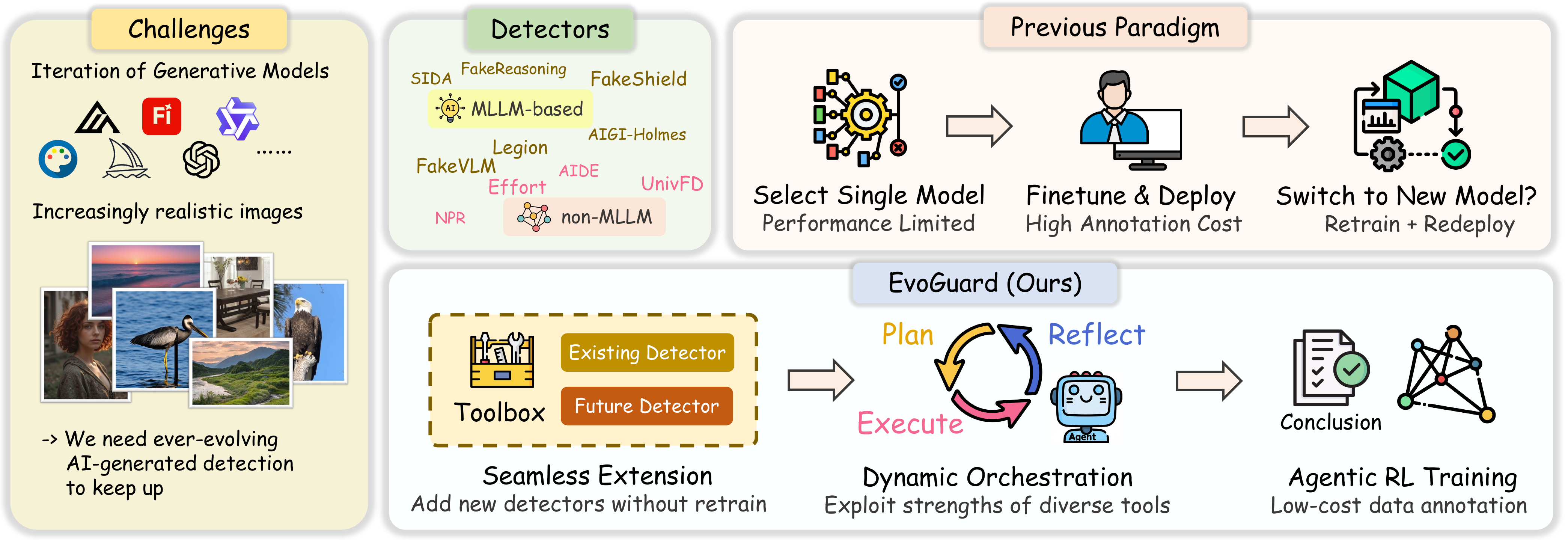}
\vspace{-2mm}
\caption{Motivation of \ouragent. As generative models rapidly evolve, AIGI detection must evolve accordingly. Prior work focuses on building ever-stronger single detectors; we instead let an MLLM agent reason over a pool of off-the-shelf detectors, cross-validating their outputs rather than routing to one or fusing all by fixed rules. This reasoning-based synthesis exploits complementary strengths from heterogeneous detectors, enables training-free extensibility, and removes the need for fine-grained annotations.}
\label{fig:motivation}
\vspace{-5mm}
\end{figure*}

\section{Introduction}

From GANs \citep{goodfellow2014generative}, Diffusion models \citep{rombach2022high} to autoregressive models \citep{fan2024fluid, liu2025infinitystar}, the rapid evolution of AI generative models has spurred unprecedented growth in the creative industry. However, the proliferation of hyper-realistic AI-Generated Images (AIGIs) has introduced severe risks of misinformation \citep{ha2024organic}, making AIGI detection a critical research area.  Existing methods include capturing generator artifacts in spatial and frequency domains \citep{cnnspot, npr}, leveraging pre-trained vision encoders \citep{effort, univfd}, and exploiting diffusion reconstruction errors \citep{dire}. However, due to the rapid iteration of generative models, image degradations on social media \citep{li2025artificial}, and sophisticated adversarial attacks \citep{tailanian2024diffusion}, existing detectors struggle in complex and ever-changing real-world environments \citep{tariang2024synthetic}. The problem of detecting AI-generated images is far from being solved \citep{aide}.

The recent emergence of Multimodal Large Language Models (MLLMs) offers a novel perspective. Unlike traditional paradigms that rely heavily on low-level features \citep{patchcraft}, MLLMs comprehend images through high-level semantics, identifying artifacts requiring global reasoning, such as logical inconsistencies or physical anomalies \citep{legion}. These semantic representations are robust to image transformations, thereby improving generalization \citep{wang2025dfbench}, while their ability to generate explanations provides a rationale for detection.  However, MLLMs are insensitive to low-level features, and often fail to capture fine-grained, pixel-level artifacts \citep{jia2024can}. Moreover, developing effective explanatory and reasoning capabilities \citep{fakevlm, huang2025so}, or aligning auxiliary vision encoder features with the LLM’s semantic space \citep{fakereasoning, zhou2025aigi}, typically requires large amounts of carefully annotated data, making it difficult to balance data quality against acquisition costs \citep{zhang2025propose}.

We argue that an effective AIGI detection system must simultaneously address three fundamental challenges:
(1) Accuracy and Generalization. The method must generalize effectively across diverse generative models and remain robust to real-world image perturbations. 
(2) Extensibility. As generative models continuously advance, detectors must be capable of swift adaptation to keep pace. 
(3) Practicality. The required training annotations should be easy to obtain.

To alleviate the aforementioned issues, we further explore MLLMs for AIGI detection, motivated by the strong decision-making and planning capabilities exhibited by modern LLMs. While prior studies largely focus on developing powerful monolithic MLLM-based detectors, the idea of tackling this task via autonomous planning within an agentic framework remains largely unexplored. An agent is an intelligent entity capable of perception, reasoning, planning, and execution, which can leverage diverse external tools to solve complex problems \citep{wang2024survey}. Agentic frameworks have been widely adopted across various domains \citep{abou2025agentic}, demonstrating remarkable adaptability, generalization, and immense potential for flexibility and extensibility. These intrinsic attributes of agents perfectly align with the core requirements of AIGI detection.

Guided by the agentic paradigm, we propose EvoGuard, which reframes AIGI detection into reasoning-based evidence synthesis through an agentic framework: 
First, we encapsulate diverse off-the-shelf detectors (both MLLM-based and non-MLLM-based) as executable tools and propose a \toolprofile~for each. 
Notably, these profiles are designed to use ambiguous, linguistic descriptors rather than exact metrics, enabling the agent to adapt better to distribution shifts.
We then devise a \mechanismnameshort~mechanism, empowering the agent to dynamically select tools based on the input image and the \toolprofile, and at each round, to reason over their heterogeneous outputs and cross-validate conflicting or low-confidence predictions before deciding whether to invoke additional tools or synthesize a final verdict.
In this way, it performs multi-round, reasoning-based evidence synthesis over heterogeneous detectors, going beyond traditional single-detector and fixed-ensemble methods.
We optimize the agent using GRPO \citep{shao2024deepseekmath}, a prevalent Agentic Reinforcement Learning (Agentic RL) algorithm, with a specially designed reward to elicit its reasoning and execution ability from low-cost binary labels alone.
Our main contributions are summarized as follows:


\begin{table*}[t]
\centering
\setlength{\tabcolsep}{4pt}
\renewcommand{\arraystretch}{1.1}

\caption{Comparison with other AIGI detection methods. 
``Could be Train-Free'' implies that it is theoretically possible to be train-free, though this has not yet been empirically validated.
* indicates available checkpoint and code for AIGI detection; these methods serve as baselines in our experiments. \textdagger~indicates methods included in the \ouragent~toolbox.}
\label{tab:methods_comparison}

\vspace{-2mm}

\resizebox{\linewidth}{!}{%
\begin{tabular}{@{}l l l c l@{}}
\toprule
\makecell[c]{Method} &
\makecell[c]{Paradigm} &
\makecell[c]{Detector Composition} &
\makecell[c]{Any-Detector Compatibility} &
\makecell[c]{Extensibility to New Detectors} \\
\midrule
Effort\textsuperscript{*}\textsuperscript{\textdagger}{\scriptsize (ICML '25)} & Single Detector & non-MLLM Detector (CLIP) & -- & -- 
\\
AIDE\textsuperscript{*}\textsuperscript{\textdagger}{\scriptsize (ICLR '25)} & Single Detector & non-MLLM Detector (CLIP + CNN) & -- & -- 
\\
MIRROR\textsuperscript{*}\textsuperscript{\textdagger}{\scriptsize (arXiv '26)} & Single Detector & non-MLLM Detector (DINO) & -- & -- 
\\
FakeReasoning\textsuperscript{*}{\scriptsize (TIP '26)} & Single Detector & MLLM-based Detector & -- & -- 
\\
LEGION {\scriptsize (ICCV '25)} & Single Detector & MLLM-based Detector & -- & -- 
\\
FakeVLM\textsuperscript{*}\textsuperscript{\textdagger} {\scriptsize (NeurIPS '25)} & Single Detector & MLLM-based Detector & -- & -- 
\\
SIDA\textsuperscript{*}{\scriptsize (CVPR '25)} & Single Detector & MLLM-based Detector + Extra Module & -- & -- 
\\
FakeShield\textsuperscript{*}{\scriptsize (ICLR '25)} & Single Detector & MLLM-based Detector + Extra Module & -- & -- 
\\
AIGI-Holmes {\scriptsize (ICCV '25)} & Single Detector & MLLM-based Detector + Extra Module & -- & -- 
\\
Forensic-MOE\textsuperscript{*}{\scriptsize (ICCV '25)} & Ensemble (MLP-based Fusion) & Multiple non-MLLM Detectors & NO & Training Required \\
E3 {\scriptsize (CVPRW '24)} & Ensemble (Transformer-based Fusion) & Multiple non-MLLM Detectors & NO & Training Required 
\\
AlignGemini {\scriptsize (arXiv '26)} & Ensemble (Fusion by OR Operation) & MLLM-based + non-MLLM Detectors & YES & Train-Free 
\\
X2-DFD {\scriptsize (NeurIPS '25)} & Pipeline (Non-MLLM outputs → MLLM) & MLLM-based + non-MLLM Detectors & YES & Could be Train-Free 
\\
\textbf{\ouragent~{\scriptsize (Ours)}} & \textbf{Agentic Framework (Dynamic)} & \textbf{Any Composition of Detectors} & \textbf{YES} & \textbf{Train-Free} \\
\bottomrule
\end{tabular}%
}

\vspace{-3mm}

\end{table*}

\begin{itemize}

\item \textbf{A novel agentic paradigm for AIGI detection.} We pioneer formulating AIGI detection as reasoning-based evidence synthesis via an agentic framework. Instead of routing to one detector or fusing outputs by fixed rules, the agent synthesizes complementary evidence from heterogeneous detectors, transcending the limits of single detectors and static ensembles.

\item \textbf{SOTA accuracy and mitigated bias.} We propose \ouragent, which leverages agent’s perception, reasoning, and planning capabilities to exploit complementary strengths of diverse tools across multiple action rounds. Extensive experiments show that \ouragent~achieves SOTA accuracy while balancing performance between positive and negative samples.

\item \textbf{Seamless extensibility and efficient training strategy.} Experiments show that \ouragent~can improve performance without retraining by plug-and-play integration of off-the-shelf new tools, offering a promising long-term solution to evolving AIGI threats. Moreover, we introduce a GRPO-based Agentic RL algorithm that trains using only a small set of low-cost binary labels, overcoming the reliance on expensive fine-grained annotations.
\end{itemize}


\section{Related Work}

\subsection{AI-Generated Image Detection}

Early AIGI detection largely relied on identifying synthetic artifacts \citep{chai2020makes}, e.g., generator fingerprints in spatial/frequency features \citep{qian2020thinking, cnnspot, frank2020leveraging} or upsampling-induced textures \citep{npr}. These cues often overfit to specific generators and are fragile to real-world degradations \citep{liu2025beyond}. More fundamentally, this paradigm treats authenticity as a sink class and ties the decision boundary to particular forgery domains \citep{univfd}, exposing significant generalization deficiencies as generative models rapidly advance.

Recent studies explored leveraging pre-trained visual encoders \citep{univfd, effort, aide, peng2025forensicssam, costanzino2025towards} such as CLIP \citep{clip}, SAM \citep{sam}, and DINO \citep{dinov2} to incorporate high-level semantics and capture artifacts requiring global reasoning \citep{legion}. 
Others compare images with authentic visual priors \citep{mirror, li2025raidx}, exploit diffusion-reconstruction errors \citep{dire, wu2025explainable}, or address bias from generation-agnostic artifacts \citep{bfree, dda, rajan2024aligned, chen2024drct, cai2025cataid}. These approaches improve individual detectors, while combining their complementary strengths remains a separate challenge.

The rapid evolution of generative models has rendered single detectors increasingly inadequate. A classic remedy is Mixtures of Experts (MoE), either by training fusion networks \citep{fang2025forensic, azizpour2024e3, liu2024mixture, kong2025moe} or directly combining detector outputs \citep{aligngemini}. Another line of work adapts detectors to new generators through additional training data \citep{epstein2023online, cf}. Compared to these methods, \ouragent~proposes the \mechanismnameshort~mechanism to dynamically invoke heterogeneous detectors and exploit their complementary strengths, enabling retraining-free, plug-and-play extensibility in response to complex, ever-changing real-world environments.

\subsection{MLLM in Visual Forensics}

MLLMs contribute semantic understanding and explanatory reasoning to visual forensics \citep{ji2025towards}. Their reliance on high-level semantics benefits robustness and generalization \citep{wang2025dfbench}, but the lack of low-level visual backbones limits their ability to capture pixel-level artifacts \citep{zhang2025propose, he2025vlforgery}, hindering forensic performance \citep{li2025fakebench, zhang2024bench, jia2024can}.

Prior work addresses this limitation by improving artifact-oriented reasoning \citep{fakevlm, tan2025semantic, huang2025thinkfake, zhang2024common, tan2025veritas, qin2025fake}, adding low-level visual features to MLLMs \citep{xu2025avatarshield, zhou2025aigi, fakereasoning, sun2025edvd, wang2025faceshield, yu2025unlocking, liu2024forgerygpt, sun2024forgerysleuth, nguyen2025prpo}, or passing specialist detector outputs as text or images \citep{x2dfd, peng2025mllm, sun2025towards}. Other methods use task-specific tokens for classification or localization \citep{sida, xu2025pixels, li2025bridging, zhang2025propose, huang2025mm, ffaa}, or build MLLM-based routers to select a suitable detector \citep{huang2025unishield}. 
Adding visual features typically requires further alignment training; existing methods rely on costly reasoning data obtained through GPT-4o \citep{huang2025so, fakeshield}, human annotation \citep{ji2025interpretable, legion}, or multiple MLLMs \citep{tan2025veritas, zhou2025aigi}. 

Unlike approaches that build a single MLLM-based detector, \ouragent~uses an MLLM agent to invoke heterogeneous detectors as tools, capturing comprehensive visual cues beyond the limits of the MLLM or any single detector. Its tailored \mechanismnameshort~mechanism enables dynamic tool scheduling and plug-and-play extensibility, offering a more intelligent and flexible alternative to conventional single-model or ensemble paradigms. Moreover, \ouragent~is trained purely with RL to elicit autonomous planning and reflection for AIGI detection without expensive fine-grained annotations. \Cref{tab:methods_comparison} compares \ouragent~with other recent methods.


\section{Methodology}

\begin{figure*}[t]
\centering
\includegraphics[scale=0.17]{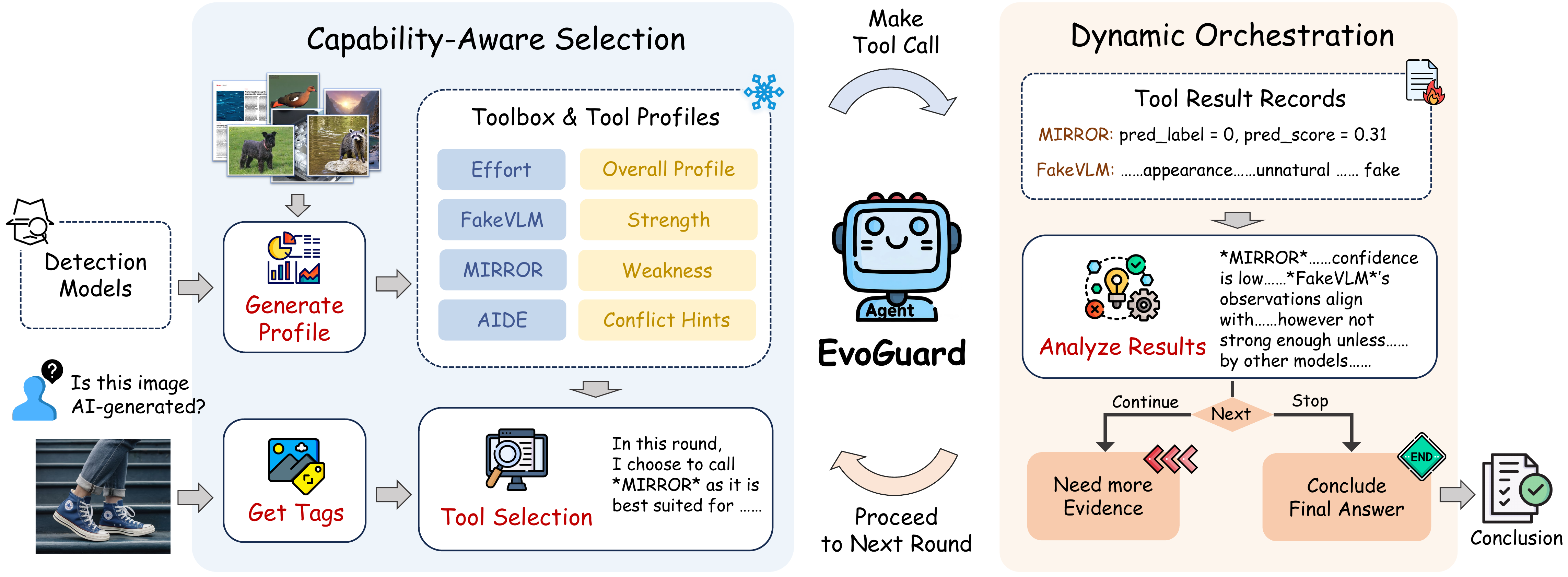}

\vspace{-2mm}

\caption{Overview of \ouragent. We wrap heterogeneous AIGI detectors as tools with capability profiles. Given a query image, the agent first performs \textbf{Capability-Aware Selection} by matching image tags with tool profiles to select suitable tools, then performs \textbf{Dynamic Orchestration} by iteratively analyzing tool outputs to decide whether to call additional tools for validation or to conclude.}
\label{fig:method}

\vspace{-4mm}

\end{figure*}

\subsection{Overview}

Practical AIGI detection must simultaneously achieve strong generalization across diverse and evolving generative models, remain extensible, and avoid reliance on costly fine-grained annotations. To address these challenges, our agentic paradigm follows three key principles:
(1) encapsulate heterogeneous detectors as tools and let an MLLM-based agent autonomously plan their use through action and reflection;
(2) leverage the agent's general understanding ability to handle heterogeneous tool outputs and support seamless integration of future tools; and
(3) train the agent via Agentic RL using only binary labels.

Following these principles, we present \ouragent. As shown in \cref{fig:method}, it consists of two key components: Capability-Aware Selection and Dynamic Orchestration. 
Capability-Aware Selection encapsulates existing detectors as tools and constructs tool profiles to select suitable tools for each input image (\cref{sec:Capability-awareSelection}). 
Dynamic Orchestration enables the agent to iteratively analyze previous tool outputs and decide whether to invoke additional tools or produce the final prediction (\cref{sec:DynamicOrchestration}). 
We train \ouragent~with GRPO using specially designed rewards and only image-level binary labels (\cref{sec:AgenticRLTraining}). 
The resulting framework also supports plug-and-play extension to future detectors without retraining (\cref{sec:ExtendtoFutureTools}).

\subsection{Capability-Aware Selection}
\label{sec:Capability-awareSelection}

The first key to the design of \ouragent~lies in the intelligent invocation of appropriate tools to obtain effective results while mitigating the negative influence of suboptimal outputs. To this end, it is essential to recognize the disparities among different tools and establish a paradigm for matching suitable tools. We therefore design a profile for each tool to characterize its capabilities, thereby representing the differences. Subsequently, we encourage the agent to select tools by integrating the input image with the tool profiles.

\textbf{Tool Profile Design.} The profile comprises two layers: \emph{structure}, which specifies the dimensions used to characterize a tool, and \emph{content}, which provides concrete descriptors within each dimension. We define four \emph{structure} dimensions: \emph{Overall Profile}, \emph{Strengths}, \emph{Weaknesses}, and \emph{Conflict Hints}. The \emph{Overall Profile} summarizes behavioral tendencies (e.g., models trained to detect generative artifacts tend to be conservative, often overlooking unseen artifacts~\citep{mirror}). \emph{Strengths} and \emph{Weaknesses} indicate data types where the tool performs particularly well or poorly. \emph{Conflict Hints} support decision-making under disagreement by emphasizing a tool's relative advantage on the given sample type rather than its overall performance. Together, these dimensions capture both holistic and fine-grained characteristics, covering absolute and relative capabilities, thereby establishing a comprehensive framework for tool description.

The \emph{content} layer selects criteria to distinguish images so that capability differences are exposed; it should be neither too coarse nor too granular. We use three tag dimensions—\emph{Subject}, \emph{Quality}, and \emph{Style}—and constrain the candidate tag set to facilitate profile generation and tool selection.

\textbf{Tool Profile Generation.} We generate tool profiles based on two principles. 1) \emph{Conciseness}. We only describe cases where metric values (accuracy, false negative rate, and false positive rate) are notably high or low on specific data types, omitting ordinary cases, as excessive context length degrades performance~\citep{du2025context} and increases inference costs. 2) \emph{Ambiguity}. Rather than specifying exact numerical values, we use linguistic expressions to convey the magnitude of metrics. This is because the training set distribution inevitably differs from real-world applications, and LLM-driven agents possess inherent reasoning capabilities; thus, maintaining ambiguity enables dynamic decision-making based on actual sample characteristics. 

In practice, we generate tool profiles using the training set. We employ the same underlying MLLM of \ouragent~to assign tags to each image, then compute each tool's metrics under each tag. Based on these metrics, we use LLM to produce an initial \toolprofile~and then manually refine it. Given a tool set $\mathbb{T} = \{t_1, t_2,\cdots,t_M\}$, we obtain the tool profile set $\mathbb{P} = \{p_1, p_2,\cdots,p_M\}$. We then use this $\mathbb{P}$ for training and inference. More details can be found in the appendix.

\textbf{Tool Selection.} To determine the tools for a given image $x$,  
\ouragent~uses its backbone MLLM to obtain its tag $g(x)$.
Then, at dialogue round $n$, \ouragent~selects a tool set $\mathbb{T}(n)$ based on the dialogue
context $c_{n-1}$ available before this round, the image $x$ and its tag $g(x)$, and the tool profile set $\mathbb{P}$:
{
\setlength{\abovedisplayskip}{3pt plus 1pt}
\setlength{\belowdisplayskip}{3pt plus 1pt}
\setlength{\abovedisplayshortskip}{0pt}
\setlength{\belowdisplayshortskip}{0pt}
\begin{equation}\label{eq:toolsel}
\mathbb{T}(n) = \mathrm{Select}(x, g(x), c_{n-1}, \mathbb{P}).
\end{equation}
}

\subsection{Dynamic Orchestration}
\label{sec:DynamicOrchestration}

Capability-Aware Selection decides which tool(s) to invoke at each round. 
We do not constrain the interaction to a single round, as tools may produce conflicting conclusions or low-confidence predictions.
We therefore introduce a Dynamic Orchestration mechanism to better leverage the agent's ability to autonomously plan actions. 
After each tool invocation, the agent reflects on the current evidence and decides whether to call more tools or to conclude. 
When tool outputs conflict or a prediction is of low confidence, the agent consults tool profiles to judge which tool is more credible on the current sample, and if still inconclusive, tends to invoke a complementary tool to gather additional evidence.

Specifically, at round $n$,  after invoking $\mathbb{T}(n)$, \ouragent~obtains the tool outputs $O_n$. Given $c_{n-1}$ and $O_n$, \ouragent~analyzes the current state and produces an analysis $h_n$ and an action $a_n \in \{\texttt{stop}, \texttt{continue}\}$. The context $c_n$ is also updated by incorporating $O_n$ and $h_n$:
{%
\setlength{\abovedisplayskip}{3pt plus 1pt}
\setlength{\belowdisplayskip}{3pt plus 1pt}
\setlength{\abovedisplayshortskip}{0pt}
\setlength{\belowdisplayshortskip}{0pt}
\begin{align}
(h_n, a_n) &= \mathrm{Analyze}(x, g(x), c_{n-1}, O_{n}, \mathbb{P}), \qquad c_n = \{c_{n-1}, O_n, h_n\}.
\end{align}%
}

Afterwards, if $a_n=\texttt{continue}$, we proceed to round $n+1$, where Eq. (1) performs tool selection using the updated context $c_n$; otherwise, the agent synthesizes the evidence accumulated across rounds, reasoning in light of the tool profiles (especially the Conflict Hints) and the image content, to resolve the conflicting predictions and produce the final prediction $\mathrm{Answer}(x)$:
{
\setlength{\abovedisplayskip}{3pt plus 1pt}
\setlength{\belowdisplayskip}{3pt plus 1pt}
\setlength{\abovedisplayshortskip}{0pt}
\setlength{\belowdisplayshortskip}{0pt}
\begin{equation}
\mathrm{Answer}(x) = \mathrm{Conclude}\!\left(x, g(x), c_n, \mathbb{P}\right).
\end{equation}%
}

Enabling reflection and additional actions essentially improves performance on complex cases via test-time scaling~\citep{snell2024scaling, xia2025mirage}. Compared with selecting a single tool~\citep{huang2025unishield} or ensembling all tools~\citep{fang2025forensic}, our approach preserves the benefits of multi-tool collaboration while mitigating interference and adaptively controlling cost based on sample-specific runtime behavior. Notably, the MLLM's general understanding capability enables the fusion of heterogeneous outputs from different detectors, including numerical scores, text, and images, which is crucial for extending to future tools. Meanwhile, such auxiliary signals beyond binary labels further help the MLLM make more informed decisions.

\subsection{Agentic RL Training}
\label{sec:AgenticRLTraining}

Under our paradigm, training focuses on learning effective tool use rather than improving a single detector. This leads us to Agentic Reinforcement Learning (Agentic RL), which reframes LLMs from passive sequence generators into autonomous, decision-making agents embedded in complex, dynamic worlds \citep{zhang2025landscape}. Since reasoning and tool-use trajectories are highly diverse and fine-grained SFT trajectories are costly to obtain, we adopt GRPO~\citep{shao2024deepseekmath}, a widely used policy optimization algorithm in Agentic RL~\citep{zhang2025landscape}, to encourage autonomous exploration. Such a paradigm aligns closely with our task setting, and reinforcement learning has also been shown to provide stronger generalization than SFT~\citep{chu2025sft}.

For reward design, in addition to the commonly used outcome-correctness $\mathrm{Reward}_{\text{acc}}$ and format-compliance $\mathrm{Reward}_{\text{fmt}}$ rewards~\citep{deepseekr1}, we introduce an explicit thinking reward $\mathrm{Reward}_{\text{analysis}}$. Motivated by prior findings that moderate reasoning can improve performance~\citep{jin2024impact, wu2025more} and our observation that insufficient deliberation degrades performance, we design $\mathrm{Reward}_{\text{analysis}}$ to penalize insufficient analysis before tool use and the final conclusion, while leaving the reasoning form unconstrained. Accordingly, our reward function $\mathrm{R}(\cdot)$ is formulated as:
{%
\setlength{\abovedisplayskip}{3pt plus 1pt minus 1pt}
\setlength{\belowdisplayskip}{3pt plus 1pt minus 1pt}
\setlength{\abovedisplayshortskip}{0pt}
\setlength{\belowdisplayshortskip}{0pt}
\setlength{\jot}{1pt}%
\begin{equation}
\label{eq:reward}
\begin{split}
\mathrm{R}(\mathrm{traj}(x))
&= \mathrm{R}_{\text{acc}}\!\big(\mathrm{gt}(x), \mathrm{Answer}(x)\big) + \mathrm{R}_{\text{fmt}}\!\big(\mathrm{traj}(x)\big)
+ \mathrm{R}_{\text{analysis}}\!\big(\mathrm{traj}(x)\big).
\end{split}
\end{equation}
}
where $\mathrm{gt}(x)$ is the ground-truth label and $\mathrm{traj}(x)$ denotes the trajectory produced by the agent for input $x$. More details about the reward are provided in the appendix.

\subsection{Extending to Future Tools}
\label{sec:ExtendtoFutureTools}

\ouragent~is inherently designed to extend to additional detectors. First, we encapsulate heterogeneous detectors as tools under a unified schema, so adding or removing tools requires only a prompt update, enabling true plug-and-play integration. Second, we leverage an MLLM to interpret tool outputs, which naturally supports diverse return formats.

For newly added tools, our experiments show that a lightweight profile describing only overall behavior (drafted from prior knowledge) is already capable of improving performance (\cref{sec:exp2}). While a comprehensive profile would evaluate the tool on a given dataset, this still avoids retraining and redeploying the agent, making the extension substantially more convenient.


\begin{table*}[t]
\centering
\caption{Performance comparison of various methods on LOKI, CommunityForensic-Eval, and Bfree-Test datasets. Method categories are indicated by distinct markers with shaded row colors: \markTool~baselines used as tools, \markBase~other baselines, and \markFw~Pretrained MLLMs. \ouragent~uses Effort, FakeVLM, and AIDE as its primary tool set (``3 tools''), with MIRROR added as a fourth tool in the extended setting. 
Each EvoGuard variant is trained with its corresponding tool set. 
The best and second-best results are highlighted in \textbf{bold} and \underline{underlined}, respectively.
B-Acc: balanced accuracy, R-Acc: real accuracy, F-Acc: fake accuracy, R/F Acc Gap: average absolute difference between R-Acc and F-Acc across datasets. 
Based on Qwen3-VL-4B-Instruct, \ouragent~improves upon its constituent tools in performance while maintaining a better real/fake balance, and achieves strong overall performance among all compared methods.
}

\vspace{-2mm}

\footnotesize
\renewcommand{\arraystretch}{1.20}
\setlength{\tabcolsep}{4pt}
\resizebox{\textwidth}{!}
{%
\begin{tabular}{>{\centering\arraybackslash}m{4.5cm}*{3}{>{\centering\arraybackslash}m{1.05cm}>{\centering\arraybackslash}m{1.05cm}c}>{\centering\arraybackslash}m{1.05cm}>{\centering\arraybackslash}m{1.05cm}c}
\toprule
\multirow{2}{*}{\centering Methods}
& \multicolumn{3}{c}{LOKI}
& \multicolumn{3}{c}{CommunityForensic-Eval}
& \multicolumn{3}{c}{Bfree-Test}
& \multicolumn{3}{c}{Average} \\
\cmidrule(lr){2-4}\cmidrule(lr){5-7}\cmidrule(lr){8-10}\cmidrule(lr){11-13}
& B-Acc & F1 & \textcolor{gray}{R-Acc / F-Acc}
& B-Acc & F1 & \textcolor{gray}{R-Acc / F-Acc}
& B-Acc & F1 & \textcolor{gray}{R-Acc / F-Acc}
& B-Acc & F1 & R/F Acc Gap \\
\hline

\rowcolor{promptable} \markTool~Effort (ICML'25)
& 0.7409 & 0.7832 & \textcolor{gray}{0.711 / 0.771}
& 0.8260 & 0.8205 & \textcolor{gray}{0.802 / 0.850}
& 0.7083 & 0.6235 & \textcolor{gray}{0.954 / 0.463}
& 0.7584 & 0.7424 & 0.200 \\

\rowcolor{promptable} \markTool~FakeVLM (NeurIPS'25)
& 0.8162 & 0.8613 & \textcolor{gray}{0.743 / 0.889}
& 0.7548 & 0.7763 & \textcolor{gray}{0.569 / 0.941}
& 0.6889 & 0.8586 & \textcolor{gray}{0.417 / 0.961}
& 0.7533 & 0.8321 & 0.354 \\

\rowcolor{promptable} \markTool~AIDE (ICLR'25)
& 0.7370 & 0.7198 & \textcolor{gray}{0.857 / 0.617}
& 0.7264 & 0.6902 & \textcolor{gray}{0.819 / 0.634}
& 0.5168 & 0.1558 & \textcolor{gray}{0.947 / 0.087}
& 0.6601 & 0.5219 & 0.428 \\

\rowcolor{mirrortable} \markTool~MIRROR (arXiv'26)
& \underline{0.8523} & \underline{0.8643} & \textcolor{gray}{0.882 / 0.822}
& \underline{0.9332} & \underline{0.9288} & \textcolor{gray}{0.975 / 0.891}
& \underline{0.9736} & \underline{0.9776} & \textcolor{gray}{0.983 / 0.964}
& \underline{0.9197} & \underline{0.9236} & \underline{0.054} \\

\rowcolor{repretable} \markBase~FakeShield (ICLR'25)
& 0.6155 & 0.7148 & \textcolor{gray}{0.476 / 0.756}
& 0.7777 & 0.7942 & \textcolor{gray}{0.607 / 0.949}
& 0.6331 & 0.7423 & \textcolor{gray}{0.550 / 0.717}
& 0.6754 & 0.7504 & 0.263 \\

\rowcolor{repretable} \markBase~SIDA (CVPR'25)
& 0.6192 & 0.5999 & \textcolor{gray}{0.731 / 0.507}
& 0.5336 & 0.3839 & \textcolor{gray}{0.768 / 0.299}
& 0.6718 & 0.5633 & \textcolor{gray}{0.940 / 0.403}
& 0.6082 & 0.5157 & 0.410 \\

\rowcolor{repretable} \markBase~FakeReasoning (TIP'26)
& 0.6478 & 0.6901 & \textcolor{gray}{0.639 / 0.657}
& 0.8038 & 0.7892 & \textcolor{gray}{0.836 / 0.772}
& 0.6684 & 0.8015 & \textcolor{gray}{0.513 / 0.824}
& 0.7067 & 0.7603 & 0.131 \\

\rowcolor{repretable} \markBase~Forensic-MOE (ICCV'25)
& 0.7492 & 0.7705 & \textcolor{gray}{0.776 / 0.723}
& 0.6867 & 0.6813 & \textcolor{gray}{0.660 / 0.713}
& 0.4940 & 0.0155 & \textcolor{gray}{0.980 / 0.008}
& 0.6433 & 0.4891 & 0.359 \\

\rowcolor{mixturetable} \markFw~Qwen3-VL-4B
& 0.534 & 0.338 & \textcolor{gray}{0.843 / 0.226}
& 0.606 & 0.555 & \textcolor{gray}{0.698 / 0.514}
& 0.694 & 0.683 & \textcolor{gray}{0.828 / 0.561}
& 0.6113 & 0.5253 & 0.356 \\

\rowcolor{mixturetable} \markFw~Qwen3-VL-8B
& 0.522 & 0.091 & \textcolor{gray}{0.997 / 0.048}
& 0.595 & 0.346 & \textcolor{gray}{0.974 / 0.215}
& 0.660 & 0.516 & \textcolor{gray}{0.967 / 0.353}
& 0.5923 & 0.3177 & 0.774 \\

\rowcolor{mixturetable} \markFw~Qwen3-VL-30B-A3B
& 0.627 & 0.534 & \textcolor{gray}{0.853 / 0.401}
& 0.770 & 0.738 & \textcolor{gray}{0.868 / 0.672}
& 0.928 & 0.925 & \textcolor{gray}{0.993 / 0.863}
& 0.7750 & 0.7323 & 0.259 \\


\rowcolor{mixturetable} \markFw~DeepSeek-V4.1-Flash (552B)
& 0.692 & 0.680 & \textcolor{gray}{0.799 / 0.585}
& 0.806 & 0.795 & \textcolor{gray}{0.815 / 0.796}
& 0.888 & 0.928 & \textcolor{gray}{0.848 / 0.928}
& 0.7953 & 0.8010 & 0.104 \\




\midrule
\textbf{\ouragent}~(\textbf{Ours}, 4B, 3 tools)
& 0.8035 & 0.8465 & \textcolor{gray}{0.747 / 0.860}
& 0.8125 & 0.8202 & \textcolor{gray}{0.680 / 0.945}
& 0.8044 & 0.8367 & \textcolor{gray}{0.832 / 0.777}
& 0.8068 & 0.8345 & 0.144 \\

\textbf{\ouragent}~(\textbf{Ours}, 4B, 4 tools)
& \textbf{0.8638} & \textbf{0.8824}& \textcolor{gray}{0.867 / 0.861}
& \textbf{0.9380} & \textbf{0.9344}& \textcolor{gray}{0.943 / 0.933}
& \textbf{0.9792} & \textbf{0.9843}& \textcolor{gray}{0.980 / 0.978}
& \textbf{0.9270} & \textbf{0.9337} & \textbf{0.006} \\
\bottomrule
\end{tabular}%
}
\label{tab:main_comparison}

\vspace{-2mm}
\end{table*}

\begin{figure*}[tbp]
  \centering
  \begin{subfigure}[b]{0.3\textwidth}
    \centering
    \includegraphics[width=\textwidth]{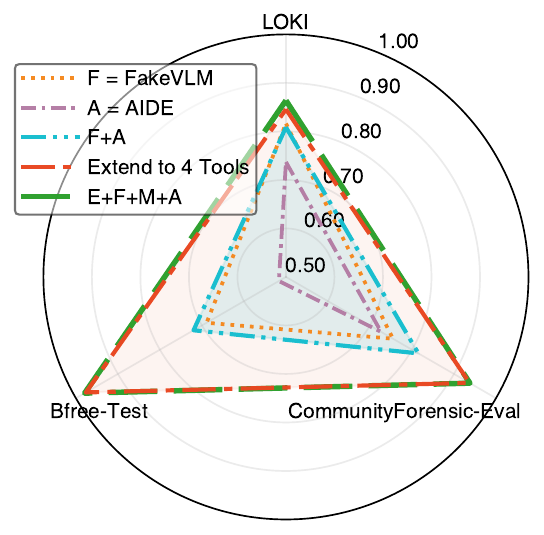}
    \caption{FakeVLM + AIDE}
    \label{fig:a}
  \end{subfigure}
  \hspace{0.01\textwidth}
  \begin{subfigure}[b]{0.3\textwidth}
    \centering
    \includegraphics[width=\textwidth]{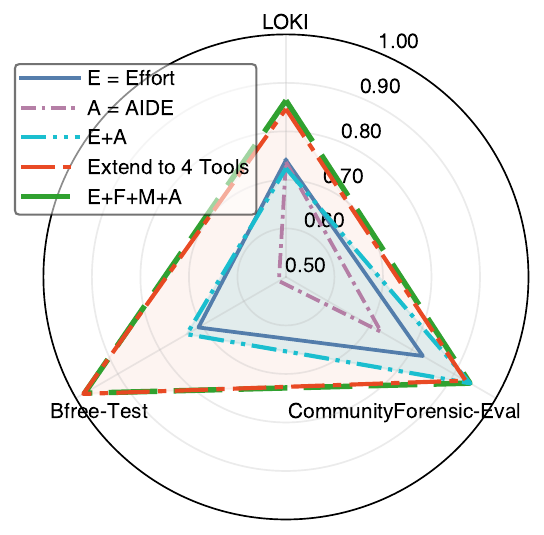}
    \caption{Effort + AIDE}
    \label{fig:b}
  \end{subfigure}
  \hspace{0.01\textwidth}
  \begin{subfigure}[b]{0.3\textwidth}
    \centering
    \includegraphics[width=\textwidth]{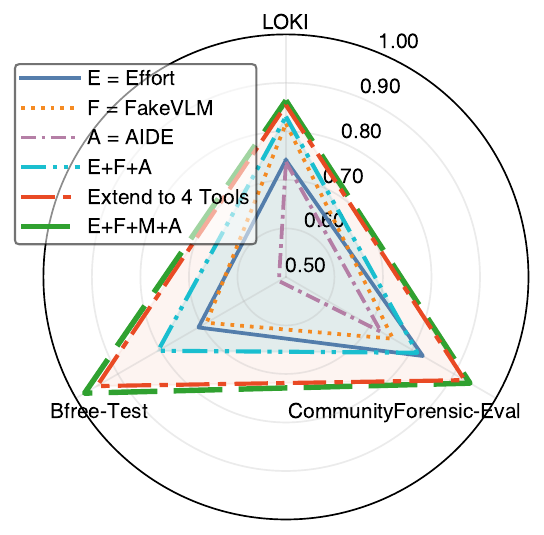}
    \caption{FakeVLM + Effort + AIDE}
    \label{fig:c}
  \end{subfigure}

\vspace{-2mm}
  
  \caption{Results of \ouragent~trained on different tool subsets then evaluated on the same tool subsets (F+A/E+A/F+E+A) and on full tool set (Extend to 4 Tools), and \ouragent~trained and evaluated on full tool set (E+F+M+A). The metric is Balanced Accuracy. Every group achieves a performance improvement through train-free extension.}
  \label{fig:extensibility}

\vspace{-4mm}

\end{figure*}

\section{Experiments}
\label{sec:experiments}

To validate the effectiveness and advantages of the proposed method, we design our experiments to answer three questions: (1) Can \ouragent~more effectively detect AIGI than existing SOTA methods (\cref{sec:exp1})? (2) Can \ouragent~be easily extended to future tools? This extensibility is crucial for rapidly responding to newly emerging real-world threats (\cref{sec:exp2}). (3) What are the individual contributions of the components in \ouragent~(\cref{sec:exp3})? We first introduce the setup of experiments, and then present results and analysis. More details on the experimental settings, computation costs, tool-call distributions, robustness evaluations, and additional results can be found in the appendix.

\subsection{Experiment Setup}

\textbf{Tool Set. }
We use three recently published SOTA methods with complementary architectures and visual cues as the primary tool set: Effort \citep{effort}, a CLIP-based detector that freezes the principal components of the feature space to prevent overfitting; FakeVLM \citep{fakevlm}, a VLM-based detector trained with fine-grained artifact-explanation supervision; and AIDE \citep{aide}, which combines DCT-based frequency and CLIP-based semantic features. We further introduce MIRROR \citep{mirror}, which measures the perplexity and residual of DINO-based \citep{dinov2} manifold-consistent references.

\textbf{Datasets. }

To simulate in-the-wild tool use while avoiding the training data of the selected tools, we construct an 8,000-image mixed training set. Using seed 42, we shuffle each source and select 1,000 real and 1,000 fake images from each of the MMFR training set, MMFR test set \citep{fakereasoning}, and DDA-COCO \citep{dda}, together with 2,000 fake images from SynthScars \citep{legion}.

For evaluation, we use three in-the-wild benchmarks: LOKI \citep{loki}, which is a recent comprehensive benchmark covering multiple image sources with broad content diversity; Bfree-Test \citep{bfree}, which contains different versions of the same images collected from social networks, making it suitable for evaluating robustness; and CommunityForensic \citep{cf}, which includes images from thousands of generators. Due to its scale, we sample a subset of it. Together, these three datasets form a strong benchmark for effectively assessing the performance of \ouragent~and baseline methods in complex and diverse in-the-wild settings.

\textbf{Implementation Details. }

\ouragent~uses Qwen3-VL-4B-Instruct \citep{qwen3vl} as the base model. We conduct GRPO training with multi-round tool calls using the AgentLoop module in the VeRL framework \citep{verl}, with the learning rate set to $1\times10^{-6}$ and the KL loss coefficient to 0.001. Only the agent is trained, while all tools remain frozen.

\textbf{Comparative Methods. }
In order to ensure a comprehensive comparison, the evaluated methods include four existing tools, other SOTA detection methods including FakeShield~\citep{fakeshield}, SIDA~\citep{sida}, FakeReasoning~\citep{fakereasoning}, and Forensic-MOE~\citep{fang2025forensic}, and pretrained MLLMs of different size, forming a diverse benchmark for comparison. For all competing methods, we use the officially released checkpoints from their authors.

\textbf{Metrics.} We employ commonly used evaluation metrics Balanced Accuracy (B-Acc) and F1 score, and accuracy on real and fake images (R-Acc / F-Acc).

\begin{table*}[t]
\caption{Ablation results on LOKI, CommunityForensic-Eval, and Bfree-test. Markers and shaded rows distinguish the ablation of Capability-Aware Selection (\markTool), training (\markTrain), and Dynamic Orchestration (\markBase), as well as comparisons with other MoE methods (\markFw). The best performance is in \textbf{bold}, while the second-best is \underline{underlined}. B-Acc: balanced accuracy. $\oplus$ denotes logical OR.}

\vspace{-2mm}

\centering
\small
\renewcommand{\arraystretch}{1.20}
\setlength{\tabcolsep}{4pt}
\resizebox{1\textwidth}{!}{%
\begin{tabular}{>{\centering\arraybackslash}m{4.8cm}*{8}{>{\centering\arraybackslash}m{1.8cm}}}
\toprule
\multirow{2}{*}{\centering Method}
& \multicolumn{2}{c}{LOKI}
& \multicolumn{2}{c}{CommunityForensic-Eval}
& \multicolumn{2}{c}{Bfree-Test}
& \multicolumn{2}{c}{Average} \\
\cmidrule(lr){2-3}\cmidrule(lr){4-5}\cmidrule(lr){6-7}\cmidrule(lr){8-9}
& B-Acc & F1 & B-Acc & F1 & B-Acc & F1 & B-Acc & F1 \\
\midrule
\ouragent~(4 tools)
& \textbf{0.8638} & \textbf{0.8824}
& \textbf{0.9380} & \textbf{0.9344}
& \textbf{0.9792} & \textbf{0.9843}
& \textbf{0.9270} & \textbf{0.9337} \\
\rowcolor{promptable} \markTool~\ouragent~(w/o Tool Profile)
& 0.7900 & 0.8392
& 0.8590 & 0.8606
& 0.9523 & 0.9696
& 0.8671 & 0.8898 \\
\rowcolor{promptable} \markTool~Match Best Tools
& 0.8150 & 0.8173
& 0.9330 & \underline{0.9302}
& 0.9736 & 0.9776
& 0.9072 & 0.9084 \\

\rowcolor{trainingtable} \markTrain~\ouragent~(w/o RL Training)
& 0.7067 & 0.8132
& 0.8201 & 0.8283
& 0.7592 & 0.8959
& 0.7620 & 0.8458 \\

\rowcolor{trainingtable} \markTrain~\ouragent~(w/o Analysis Reward)
& 0.8505 & \underline{0.8684}
& 0.8933 & 0.8883
& 0.9643 & 0.9782
& 0.9027 & 0.9116 \\

\rowcolor{repretable} \markBase~\ouragent~(select one tool)
& \underline{0.8513} & 0.8639
& \underline{0.9332} & 0.9287
& \underline{0.9755} & \underline{0.9786}
& \underline{0.9200} & \underline{0.9237} \\
\rowcolor{repretable} \markBase~\ouragent~(invoke all tools)
& 0.8417 & 0.8603
& 0.9138 & 0.9098
& 0.9408 & 0.9634
& 0.8988 & 0.9112 \\
\rowcolor{repretable} \markBase~\ouragent~(one round)
& 0.8316 & 0.8509
& 0.9181 & 0.9139
& 0.9704 & 0.9780
& 0.9067 & 0.9143 \\
\rowcolor{mixturetable} \markFw~MoE by Gate Network
& 0.7419 & 0.8229
& 0.9113 & 0.9072
& 0.9660 & 0.9763
& 0.8731 & 0.9021 \\
\rowcolor{mixturetable} \markFw~Weighted Majority Vote
& 0.8496 & 0.8598
& 0.9227 & 0.9183
& 0.9601 & 0.9645
& 0.9108 & 0.9142 \\
\rowcolor{mixturetable} \markFw~Confidence Averaging
& 0.8462 & 0.8563
& 0.8828 & 0.8805
& 0.9586 & 0.9628
& 0.8959 & 0.8999 \\

\rowcolor{mixturetable} \markFw~Calibrated Score Average
& 0.7893 & 0.8567
& 0.8016 & 0.8175
& 0.8854 & 0.9433
& 0.8254 & 0.8725 \\
\rowcolor{mixturetable} \markFw~Calibrated Weighted Vote
& 0.7988 & 0.8588
& 0.8144 & 0.8268
& 0.9158 & 0.9588
& 0.8430 & 0.8815 \\

\rowcolor{mixturetable} \markFw~FakeVLM $\oplus$ MIRROR
& 0.8003 & 0.8671
& 0.7635 & 0.7883
& 0.7047 & 0.8744
& 0.7562 & 0.8433 \\
\bottomrule
\end{tabular}%
}
\label{tab:ablation}

\vspace{-4mm}

\end{table*}

\subsection{Performance Comparison}\label{sec:exp1}

\textbf{Generalization on In-the-Wild Benchmarks.}
\Cref{tab:main_comparison} presents the performance comparison between \ouragent~and comparative methods. With its primary three tools, \ouragent~improves average B-Acc by nearly 5 points over the best individual tool and exceeds every other baseline except MIRROR, including much larger pretrained MLLMs. Although MIRROR already achieves 91.97\% average B-Acc, \ouragent~further improves performance across all three benchmarks with MIRROR in its tool set. These results demonstrate that \ouragent~effectively integrates different detectors and leverages their complementary strengths to achieve gains beyond any individual tool.

\textbf{Mitigating Detector Bias.} \ouragent~consistently reduces the average R/F Acc Gap of its constituent tools in both settings.  While many existing detectors exhibit severe class bias, \ouragent~narrows the real/fake accuracy gap by 28--66\% in the three-tool setting, and by 9--70$\times$ after adding MIRROR (\cref{tab:main_comparison}). This shows that \mechanismnameshort~can exploit complementary evidence to mitigate detector bias, with the degree of mitigation depending on the available tools. Such biases may arise from spurious features \citep{dda, bfree} or limitations of existing detection paradigms \citep{mirror}

\textbf{Fusion of Heterogeneous Detectors.} Among the 4 selected tools, FakeVLM returns only textual outputs, while the other three return confidence scores. Leveraging the general reasoning ability of MLLMs, \ouragent~can reason over outputs in different formats, which is also crucial for future extensibility. Detailed examples are provided in the appendix.

\subsection{Extensibility to Future Tools}\label{sec:exp2}

The rapid evolution of generative models and forgery methods requires detectors to continuously evolve as well. A detector that can improve efficiently and continuously would be highly valuable. An important design goal of \ouragent~is to scale to more tools without training. In this way, when new detection tools emerge in the future, \ouragent~can incorporate them directly, thereby improving its ability to handle novel forgeries. 

To investigate the effectiveness of this design, we train \ouragent~on a subset of tools and evaluate whether introducing held-out tools only at inference time improves performance. For newly introduced tools, we supply only the \emph{Overall Profile}, mimicking a realistic setting where only coarse information about a new tool is available. Since the number of possible subsets is large, we construct representative 3 tool combinations according to two principles: (1) selecting heterogeneous detectors to exploit complementarity, and (2) simulating the introduction of newer, stronger detectors. The results are shown in~\cref{fig:extensibility}.

\textbf{Train-free Extensibility.} As shown, agents trained on each tool subset benefit from the addition of previously unseen tools at test time. This train-free tool extension yields substantial gains, bringing performance close to that of an agent trained with the full tool set. These results suggest that \ouragent~effectively learns to leverage diverse tools during training and can generalize this capability to newly introduced tools.

\subsection{Ablation \& Mechanism Analysis}
\label{sec:exp3}

We further analyze the core design of \ouragent~through multi-dimensional ablations with the full tool set and comparisons with routing-based and other ensemble methods, as summarized in~\cref{tab:ablation}.

\textbf{Capability-Aware Selection and Training.} Removing the Tool Profile reduces performance across all three benchmarks, showing the value of explicit guidance about tool capabilities. We also test using an image's three-dimensional tags to match three best-performing tools and taking a majority vote, which seems more precise, yet underperforms \ouragent, likely due to train--test data shift. This suggests that the deliberate ambiguity in \toolprofile~helps it generalize.
Then, without RL training or removing the analysis reward, performance also drops even though the base model and prompts remain the same. These results show that capability guidance and training both contribute to effective tool use.

\textbf{Dynamic Orchestration and Other MoE Methods.} Restricting \ouragent~to selecting one tool, invoking all tools, or using only one round reduces performance in every case, supporting the benefit of adapting both tool selection and the inference process to the image. Meanwhile, \ouragent~outperforms all compared methods on both B-Acc and F1 across the three benchmarks, including learned gating, voting, score averaging, and logical OR. The results demonstrate the advantage of reasoning over tool outputs and coordinating their use according to the input. Implementation details of the compared methods and illustrative cases are provided in the appendix.


\section{Discussion and Conclusion}

\textbf{Conclusion.} We present \ouragent, an agentic framework that recasts AIGI detection as reasoning-based evidence synthesis over a pool of heterogeneous detectors. Going beyond single detectors and conventional ensembles, \ouragent~reasons over their heterogeneous outputs, cross-validating conflicting or low-confidence signals before concluding. Trained via Agentic RL with only low-cost binary labels, it achieves SOTA accuracy across diverse benchmarks while markedly mitigating the real/fake class bias. Most importantly, it supports plug-and-play integration of new tools without retraining, enabling continual evolution against emerging threats.

\textbf{Discussion.} While \ouragent~offers a practical and sustainable solution, it also introduces additional MLLM overhead and remains bounded by its underlying toolset: if all tools fail, its reasoning may also be misled.

\textbf{Future Work.} Future directions include (1) developing task-specific small MLLMs via distillation~\citep{xu2024survey}, (2) continually updating tools and profiles with real-world samples, and (3) extending \ouragent~to broader visual forensics tasks to explore the potential of this paradigm.

\bibliography{references}
\bibliographystyle{arxiv}

\end{document}